\documentclass[10pt,twocolumn,letterpaper]{article}

\usepackage{cvpr}              
\definecolor{cvprblue}{rgb}{0.21,0.49,0.74}

\usepackage[pagebackref,breaklinks,colorlinks,allcolors=cvprblue]{hyperref}

\usepackage[table,xcdraw]{xcolor}
\usepackage[most]{tcolorbox} 
\newtcolorbox[number within=chapter,]{prompt}[3][]{
arc=5mm,
lower separated=false,
fonttitle=\bfseries,
colbacktitle=blue!5,
coltitle=blue!50!black,
enhanced,
attach boxed title to top left={xshift=0.5cm,
        yshift=-2mm},
colframe=blue!50!black,
colback=blue!10,
overlay={
\node[draw=blue!50!black,thick,
fill= blue!10,rounded corners=1mm, 
yshift=0pt, 
xshift=-0.5cm, 
left, 
text=blue!50!black,
anchor=east,
font=\bfseries] 
at (frame.north east) {#3};},
overlay={
\node[draw=blue!50!black,thick,
fill= yellow!30,rounded corners=1mm, 
yshift=+1.2mm, 
xshift=-0.5cm, 
left, 
text=blue!50!black,
anchor=east,
font=\bfseries] 
at (frame.north east) {#3};},
title=#2,#1,breakable}

\def\paperID{*****} 
\def\confName{CVPR}
\def\confYear{2026}

\title{Evaluating Multimodal Large Language Models\\ for Heterogeneous Face Recognition}

\author{\vspace{3pt}Hatef Otroshi Shahreza, Anjith George, Sébastien Marcel \\
Idiap Research Institute, Switzerland\\
{\tt\small \{hatef.otroshi,anjith.george,sebastien.marcel\}@idiap.ch}\\
}

\begin{document}
\maketitle

\begin{abstract}
Multimodal Large Language Models (MLLMs) have recently demonstrated strong performance on a wide range of vision-language tasks, raising interest in their potential use for biometric applications. In this paper, we conduct a systematic evaluation of state-of-the-art MLLMs for \emph{heterogeneous face recognition} (HFR), where enrollment and probe images are from different sensing modalities, including visual (VIS), near infrared (NIR), short-wave infrared (SWIR), and thermal camera. We benchmark multiple open-source MLLMs across several cross-modality scenarios, including VIS-NIR, VIS-SWIR, and VIS-THERMAL face recognition. The recognition performance of MLLMs is evaluated using biometric protocols and based on different metrics, including Acquire Rate, Equal Error Rate (EER), and True Accept Rate (TAR). Our results reveal substantial performance gaps between MLLMs and classical face recognition systems, particularly under challenging cross-spectral conditions, in spite of recent advances in MLLMs. Our findings highlight  the limitations of current MLLMs for HFR and also the importance of rigorous biometric evaluation when considering their deployment in face recognition systems.
\end{abstract}

\section{Introduction}

Face recognition has achieved remarkable progress over the past decade, driven largely by deep convolutional and transformer-based models trained on large-scale visual datasets~\cite{deng2019arcface,george2023edgeface}. Existing face recognition systems often perform reliably in \emph{homogeneous} settings, where enrollment and probe images are captured under the same modality. However, the  performance of such models degrade significantly in \emph{heterogeneous face recognition} (HFR) scenarios, where enrollment and probe images are from different sensing modalities. Heterogeneous face recognition is particularly important, where we do not have enough data to train a model for recognition, such as new sensing cameras or special environmental conditions.
Examples of such applications include surveillance, border control,  low-light environments, etc.

\begin{figure}
    \centering
    \includegraphics[width=1\linewidth,trim={1.275cm 1.10cm 0.75cm 0.5cm},clip]{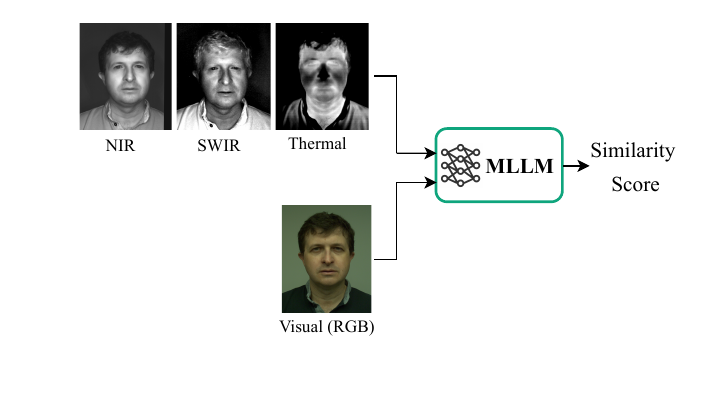}
    \caption{Heterogeneous face recognition using Multimodal Large Language Models (MLLMs)}
    \label{fig:hfr-mllm}
\end{figure}

In recent years, Multimodal Large Language Models (MLLMs) have emerged as a popular and powerful subset of foundation models that are able to process visual and textual inputs. Recent MLLMs demonstrate significant performance in  zero-shot and few-shot scenarios on different vision-language tasks, including image captioning, visual question answering, and multimodal reasoning. These capabilities have led to growing interest in whether MLLMs can be directly applied to different vision tasks, such as face recognition, and potentially offering a  unified framework that provides perception and reasoning~\cite{shahreza2025foundation,zhang2024vision,awais2025foundation}.

However, face recognition, particularly in heterogeneous settings, differs fundamentally from many vision-language tasks. Biometric recognition requires fine-grained identity discrimination, robustness to sensing variations, and adherence to strict evaluation protocols. HFR is specially a challenging task, since data from two different sensor types are compared and require strong perception and reasoning to match both data types. 
While several recent studies have explored the use of large vision or multimodal models for face-related tasks~\cite{shahreza2025foundation,sony2025benchmarking,hassanpour2024chatgpt,narayan2025facexbench,shahreza2025facellm}, their effectiveness for HFR remains unexplored. In particular, it is unclear how well MLLMs can generalize across sensing modalities and how their performance compares to established face recognition baselines.

In this paper, we present an empirical evaluation of state-of-the-art MLLMs for heterogeneous face recognition. We focus on assessing existing open-source MLLMs under standardized biometric evaluation protocols. We consider different sensing modalities, including visual (VIS), near infrared (NIR), 
short-wave infrared (SWIR), and thermal camera, and evaluate multiple MLLMs across diverse cross-spectral scenarios, including VIS-NIR, VIS-SWIR, and VIS-THERMAL (Fig.~\ref{fig:hfr-mllm}). We also compare the performance of MLLMs with face recognition models as reference baselines. 
To our knowledge, this paper presents the first evaluation of MLLMs for heterogeneous face recognition.

While MLLMs are pretrained on large amount of data, the vast majority of available image data are visual (RGB) images.  Our experimental results indicate that, despite their strong multimodal reasoning capabilities, current MLLMs struggle to match the performance of dedicated face recognition models in heterogeneous settings. Performance degradation is particularly pronounced in challenging cross-spectral scenarios, underscoring the gap between general-purpose multimodal understanding and biometric identity recognition. Our findings provide insights into the current limitations of MLLMs for domain change, such as in HFR, and outline important considerations for future research at the intersection of foundation models and biometric recognition.

The remainder of the paper is organized as follows. In Section~\ref{sec:related_work}, we review related work in the literature. In Section~\ref{sec:MLLM_HFR}, we explain our benchmark and present our results in Section~\ref{sec:experimenets}. Finally, the paper is concluded in Section~\ref{sec:conclusion}.

\section{Related Work}\label{sec:related_work}
\subsection{Heterogeneous Face Recognition}
The main challenge in HFR is the ``domain gap'', the significant differences in appearance between images taken with different sensors. This difference makes it hard to accurately match faces and reduces recognition performance. Several approaches have been proposed in the literature to address this challenge.

Common-space projection methods aim to reduce the domain gap by learning mappings that project heterogeneous face representations into a shared subspace \cite{kan2015multi,he2017learning}. Early studies relied on linear techniques such as CCA \cite{yi2007face}, PLS \cite{sharma2011bypassing}, and regression-based mappings \cite{lei2009coupled,lei2012coupled} to align facial features across modalities like NIR and VIS, while discriminative and prototype-based subspace learning further improved cross-modal recognition \cite{lin2006inter,klare2012heterogeneous}. More recent deep learning approaches utilize this concept into CNN architectures, for instance through domain-specific units that adapt low-level features across sensing modalities \cite{de2018heterogeneous}, or discriminator-based frameworks that explicitly learn cross-modality relationships using metric learning objectives \cite{cheema2022heterogeneous}. 

Invariant feature–based methods focus on extracting modality-agnostic representations that remain stable across heterogeneous sensing conditions enabling cross-modal face matching. Early approaches emphasized handcrafted descriptors, such as DoG type representations combined with multi-scale LBP features to capture structural consistency across domains \cite{liao2009heterogeneous,liaoLearningMultiscaleBlock2007}. Local feature–driven strategies further incorporated SIFT and MLBP descriptors within discriminative subspace learning frameworks to address modality gap~\cite{klare2010matching,lowe2004distinctive,ojala2002multiresolution}. Information-theoretic methods such as coupled encoding schemes designed to maximize mutual information between heterogeneous modalities \cite{zhang2011coupled} were also used for HFR. More recently, CNN-based approaches have emerged to learn invariant facial representations directly from data, demonstrating improved generalization across modalities in HFR tasks \cite{he2017learning,he2018wasserstein}. Recently, xEdgeFace \cite{xedgeface} introduced a lightweight HFR framework that adapts a pre-trained CNN-Transformer face recognition backbone \cite{george2023edgeface} to cross-modal settings by selectively updating Layer Normalization and early layers within a contrastive self-distillation setting. This strategy effectively aligns heterogeneous modalities while preserving RGB recognition performance, achieving state-of-the-art results across multiple HFR benchmarks with very low computational overhead.

Synthesis-based approaches for HFR aim to reduce modality gap by transforming images from the target domain into the source domain (typically visible spectrum images), allowing recognition with conventional face recognition models \cite{tang2003face,fu2021dvg}. Early methods relied on patch-based or manifold learning techniques for cross-modal image generation \cite{wang2008face,liu2005nonlinear}, while recent advances leverage GAN-based frameworks to synthesize identity-preserving visible faces from heterogeneous modalities, significantly improving recognition performance \cite{zhuUnpairedImagetoImageTranslation2017,zhang2017generative,fu2021dvg,george2023bridging}. However, these methods are typically computationally heavy due to the inclusion of an additional image generation stage.

\subsection{MLLMs and Biometrics}
In recent years, several papers have explored applications of MLLMs  for biometrics,  including biometric recognition, attribute analysis, forgery detection,  presentation attack detection, etc. A recent survey \cite{shahreza2025foundation}  review how MLLMs are being applied in biometrics recognition and security. 

Early studies investigated the use of pretrained MLLMs, such as ChatGPT~\cite{hurst2024gpt}, for biometric recognition (e.g., face~\cite{hassanpour2024chatgpt} or iris recognition~\cite{farmanifard2024chatgpt}) and attribute prediction. %
Jia \textit{et al.} \cite{jia2024can} evaluated the performance of ChatGPT for zero-shot face deepfake detection. 
Komaty \textit{et al.} ~\cite{komaty2025exploring} explored in-context learning using ChatGPT~\cite{hurst2024gpt} for face presentation attack detection.   
Shi \textit{et al.} \cite{shi2024shield} used chain-of-thoughts prompting with  ChatGPT and Gemini for face deepfake detection and presentation attack detection  tasks. 
Sony \textit{et al.} \cite{sony2025foundation} explored foundation models (such as CLIP, BLIP, etc.) for face recognition. 

Several benchmarks were also proposed for evaluating MLLMs for face processing tasks. 
Gavas \textit{et al.} proposed FPBench for benchmarking MLLMs for fingerprint recognition. 
Wang \textit{et al.} proposed FaceBench~\cite{wang2025facebench} as a visual question-answering benchmark for facial attribute prediction with MLLMs. Similarly, other benchmarks, such as FaceXBench~\cite{narayan2025facexbench}, FaceRecBench~\cite{shahreza2025benchmarking} and Face-Human-Bench~\cite{qin2025face} were proposed to benchmark MLLMs across various face processing tasks, including face recognition, facial expression recognition, attribute prediction,  anti-spoofing, etc. 
FaceXBench~\cite{narayan2025facexbench} and FaceRecBench~\cite{shahreza2025benchmarking} include face recognition benchmark based on face recognition datasets. FaceXBench~\cite{narayan2025facexbench} used multiple-choice questions for evaluating the performance of MLLMs, while FaceRecBench~\cite{shahreza2025benchmarking} used Yes/No questions for matching pairs of images. 
In this paper, we focus on heterogeneous face recognition and benchmark MLLMs and compare them with typical face recognition models with similar protocols. To our knwoledge, this paper is the first study on the application of MLLMs for heterogeneous face recognition.

\section{Evaluating MLLMs for Heterogeneous Face Recognition}\label{sec:MLLM_HFR}
In this section, we describe how to use Multimodal Large Language Models (MLLMs) for heterogeneous face recognition (HFR) and our evaluation pipeline for existing models under standardized biometric protocols.

\subsection{Heterogeneous Face Recognition}
Heterogeneous face recognition involves determining whether two face images belong to the same identity when the images are acquired using different sensing modalities. Let $(x_e, x_p)$ denote an enrollment-probe image pair, where $x_e$ and $x_p$ may originate from different modalities (e.g., VIS-NIR, VIS-SWIR, VIS-THERMAL). The task is to output a similarity score or decision indicating whether the given pair of images corresponds to the same identity.

\subsection{Model Inference and Prompting}
We consider each MLLM in a zero-shot setting using a fixed prompt template designed to elicit identity comparison between two face images. As a system prompt, we first describe the task for the MLLM: \textit{"You are an AI assistant specialised in heterogeneous face recognition."} Then, the user prompt instructs the model to assess whether the two images (from two different modalities) depict the same individual, without providing any identity labels:

\begin{prompt}[]{Prompt}{User}
I give you two face images taken under \texttt{modality\_1} and \texttt{modality\_2}. On a scale from 0 to 100, how likely (as a single number) are these two faces of the same person? Only output a single number (no other text)."
\end{prompt}

As we discuss later in our experiment setup, \texttt{modality\_1} and \texttt{modality\_2} can be visual, NIR, SWIR, or thermal.

Based on the prompt, the output of MLLM is expected to be a single number as similarity score of two images. This score is then used to construct genuine and impostor score distributions following standard biometric evaluation practices. As baselines, we use a classical  face recognition model and a model trained for HFR setting. We benchmark these baselines with the same protocols by extracting embeddings and calculating cosine similarity of embeddings.

\subsection{Performance Metrics}

We evaluate the recognition performance is evaluated using standard biometric metrics. We report the Equal Error Rate (EER), defined as the operating point where the False Accept Rate (FAR) equals the False Reject Rate (FRR). In addition, we report the True Accept Rate (TAR) at a fixed FAR of 1\%, which reflects performance under a security-constrained operating condition. 

While we expect a similarity score for each pair of images, not all models successfully process every image pair. We explicitly account for such cases by reporting the \emph{Acquire Rate (AR)}, defined as the proportion of image pairs for which a valid score is produced. Image pairs for which a model fails to return a usable output are counted as failures to acquire and excluded from subsequent matching performance metrics.

\section{Experiments}\label{sec:experimenets}
\subsection{Experimental Setup}
\subsubsection{Dataset}
In our experiments, we use three different datasets and evaluate HFR with  color (RGB), near-infrared (NIR), short-wave infrared (SWIR), and thermal images. For each dataset, we consider 1,000 random positive pairs (from different modalities) and 1,000 random negative pairs.
\paragraph{MCXFace Dataset:}
The MCXFace dataset \cite{george2022prepended} consists of  multi-channel image samples for 51 subjects. For each subject color (RGB), thermal, near-infrared (850 nm), short-wave infrared (1300 nm), depth, stereo depth, and depth estimated from RGB images are collected in different channels under three different sessions and various illumination conditions. We use color (RGB), thermal, near-infrared images of this dataset for evaluation of HFR under VIS-NIR and VIS-THERMAL conditions.

\paragraph{CASIA NIR-VIS 2.0 Dataset:}
The CASIA NIR-VIS 2.0 Face dataset \cite{li2013casia} includes images of 725 individuals captured under real-world, unconstrained acquisition conditions with variations in pose, expression, and illumination for both visible spectrum and near-infrared
(NIR). Each subject in the dataset 
is captured in both NIR and VIS modalities across multiple imaging sessions (1-22 visible spectrum photos and 5-50 NIR photos.)

\paragraph{Polathermal Dataset:}
The Polathermal dataset \cite{hu2016polarimetric}, collected by the U.S. Army Research Laboratory (ARL), consists of both polarimetric  long-wave infrared (LWIR) imagery  with Stokes parameter representations and color images. 
The dataset includes 60 subjects acquired under varying distances. 
In our experiments, we use the thermal and color images from this dataset. 
We use 25 identities for the training set (of baseline HFR)
and the remaining 35 identities for the test set. 

\subsubsection{Pretrained MLLMs}
We consider multiple open-source MLLMs with different sizes and structures, and evaluate their performance for HFR:

\paragraph{Gemma 3:}
Gemma 3 \cite{team2025gemma} is a MLLM developed by Google from the Gemma family \cite{team2024gemma} of lightweight open large language models, ranging in scale from 1 to 27 billion parameters, that processes both text and images and generates text output. These models are able to handle long context and can support up to a 128 k-token context window. This is achieved by increasing the ratio of local to global attention layers, while keeping the span on local attention short. The Gemma 3 models are trained with distillation and achieve superior performance to Gemma 2 for both pre-trained and instruction finetuned versions. We use Gemma-3-4B-it and Gemma-3-27B-it models in our experiments.

\paragraph{LLaVA:}
LLaVA \cite{liu2024improved} is a vision-language model that extends LLMs with visual understanding capabilities through multi-stage training which combines vision encoders with instruction-tuned language models. It leverages image-text instruction data and alignment techniques to enable processing visual input along with textual input, achieving significant performance on a wide range of multimodal benchmarks while being lightweight and efficient. LLaVA-1.5 achieved significant improvement in visual question answering, image description, and general-purpose multimodal interaction, making it a powerful model for downstream vision-language applications. We use LLaVA-1.5-7B model in our experiments.

\paragraph{Mistral}
Mistral-3.1 is an open-source MLLM developed by Mistral AI that has significant performance with 24 billion parameters. It builds on its predecessor by integrating strong text and vision understanding, supporting a long context window of up to 128 k tokens, and delivering competitive performance across instruction following, conversational assistance, multilingual tasks, code generation, and long-document reasoning. Mistral-3.1 achieved fast inference speeds and can be deployed on consumer-grade hardware while maintaining high capability across diverse generative and analytic workloads.
We use Mistral-3.1-24B model\footnote{Available at \href{https://huggingface.co/mistralai/Mistral-Small-3.1-24B-Instruct-2503}{https://huggingface.co/mistralai/Mistral-Small-3.1-24B-Instruct-2503}} in our experiments.

\paragraph{Aya-Vision:}
The Aya-Vision family \cite{dash2025aya} comprises open-weights MLLMs developed by Cohere that integrate advanced visual understanding with multilingual text capabilities across 23 languages. Built by combining a  multilingual language backbone with a state-of-the-art vision encoder, Aya-Vision models excel at several vision-language tasks, such as image captioning, optical character recognition, visual reasoning, and visual question answering. These models achieve competitive or leading performance on multilingual multimodal benchmarks. We use 
Aya-Vision-8B  
and Aya-Vision-32B 
models in our experiments.

\paragraph{InternVL3:} 
InternVL3~\cite{zhu2025internvl3} is a  series of open-source MLLM that unifies vision and language capabilities through a native multimodal pre-training paradigm, jointly learning from both multimodal and pure-text data in a single stage. It adopts a ViT-MLP-LLM architecture with a new variable visual position encoding (V2PE) to handle extended multimodal contexts effectively. InternVL3 models achieved competitive performance on a wide range of multimodal reasoning, visual comprehension, and language tasks. 
We use InternVL3-9B in our experiments.

\paragraph{Qwen3-VL}
Qwen3-VL~\cite{bai2025qwen3vltechnicalreport} is a  family of open-weight MLLMs  developed by Alibaba that has multimodal understanding across text, images, and video. It combines a vision encoder with a LLM decoder, enabling understanding and reasoning over visual and textual inputs with a native long context of up to 256 K tokens (expandable toward 1 M) for processing long documents and videos. Qwen3-VL achieved competitive performance across diverse benchmarks, such as visual question answering, OCR, spatial and 2D/3D reasoning, etc. We use 
Qwen3-VL-4B-Instruct and 
Qwen3-VL-8B-Instruct 
in our experiments.

\begin{table*}[t]
\centering
\setlength{\tabcolsep}{6.5pt}
\caption{Heterogeneous Face Recognition (HFR) performance across modalities in the MCXFace dataset. AR denotes Acquire Rate. TAR is reported at FAR=1\%. }
\label{tab:results-mcxface}
\begin{tabular}{l|ccc|ccc|ccc}
\toprule
 & \multicolumn{9}{c}{\textbf{Modality}} \\
\cmidrule(lr){2-10}
\textbf{Model}
& \multicolumn{3}{c}{VIS--NIR}
& \multicolumn{3}{c}{VIS--SWIR}
& \multicolumn{3}{c}{VIS--THERMAL} \\
\cmidrule(lr){2-4}\cmidrule(lr){5-7}\cmidrule(lr){8-10}
 & AR$\uparrow$ & EER$\downarrow$  & TAR$\uparrow$ 
 & AR$\uparrow$ & EER$\downarrow$  & TAR$\uparrow$ 
 & AR$\uparrow$ & EER$\downarrow$  & TAR$\uparrow$  \\
\midrule
        \rowcolor[HTML]{eafaf1}
         \multicolumn{10}{c}{\textbf{Open source MLLMs}} 
        \\
Gemma-3-4B-it  & 100\% & 17.40\% & 0.00\% & 100\% & 32.00\% & 0.00\% & 100\% & 52.00\% & 0.00\% \\ 
Gemma-3-27B-it  & 100\% & 5.80\% & 35.60\% & 100\% & 21.00\% & \textbf{14.70}\% & 100\% & 38.70\% & 0.80\% \\ 
LLaVA-1.5-7B-hf  & 100\% & 50.20\% & 0.20\% & 100\% & 52.36\% & 0.40\% & 100\% & 47.95\% & 0.40\% \\ 
Mistral-3.1-24B-Instruct  & 94\% & 36.88\% & 1.79\% & 93\% & 40.81\% & 2.89\% & 89\% & 45.42\% & 0.11\% \\ 
Aya-Vision-8B  & 100\% & 29.05\% & 0.50\% & 99\% & 58.69\% & 0.10\% & 100\% & 50.45\% & 0.10\% \\ 
Aya-Vision-32B  & 90\% & 38.60\% & 0.22\% & 91\% & 47.29\% & 2.09\% & 89\% & 50.74\% & 1.00\% \\ 
InternVL3-9B  & 100\% & 7.93\% & 21.94\% & 100\% & 25.20\% & 5.53\% & 100\% & 43.22\% & \textbf{9.44}\% \\ 
Qwen3-VL-4B-Instruct  & 100\% & 3.80\% & 84.20\% & 100\% & 52.40\% & 5.80\% & 100\% & \textbf{16.30}\% & 9.00\% \\ 
Qwen3-VL-8B-Instruct  & 100\% & \textbf{2.90}\% & \textbf{85.70}\% & 100\% & \textbf{12.10}\% & 13.00\% & 100\% & 28.80\% & 4.00\% \\ 
\midrule
        \rowcolor[HTML]{FFF7ED}
         \multicolumn{10}{c}{\textbf{Face Recognition Models}} 
        \\
xEdgeFace & 100\% & 0.00\% & 100\% & 100\% & 0.20\% & 99.80\% & 100\% & 5.80\% & 82.00\% \\ 
EdgeFace & 100\% & 0.00\% & 100\% & 100\% & 0.70\% & 99.50\% & 100\% & 22.50\% & 30.30\% \\

\bottomrule
\end{tabular}
\end{table*}

\begin{table}[t]
\centering
\small
\setlength{\tabcolsep}{3pt}
\caption{VIS--NIR heterogeneous face recognition results on the CASIA datset.}
\label{tab:results-casia}
\begin{tabular}{lccc}
\toprule
\textbf{Model} & \scalebox{0.95}{\textbf{Acquire Rate}} & \textbf{EER} & \scalebox{0.95}{\textbf{TAR@FAR=1\%}} \\
\midrule
        \rowcolor[HTML]{eafaf1}
         \multicolumn{4}{c}{\textbf{Open source MLLMs}} 
        \\
Qwen3-VL-8B-\scalebox{0.925}{Instruct}            & 100\% & 2.90\%  & 35.10\% \\
InternVL3-9B                    & 100\% & 16.82\% & 23.40\% \\
Gemma-3-27B-it                   & 100\% & 21.56\% & 0.00\%  \\
\midrule
        \rowcolor[HTML]{FFF7ED}
         \multicolumn{4}{c}{\textbf{Face Recognition Models}} 
        \\
xEdgeFace                       & 100\% & 0.00\%  & 100\%  \\
EdgeFace                        & 100\% & 0.10\%  & 99.84\% \\
\bottomrule
\end{tabular}
\end{table}

\begin{table}[t]
\centering
\small
\setlength{\tabcolsep}{3pt}
\caption{VIS--THERMAL Heterogeneous Face Recognition results on the Polarimetric dataset.}
\label{tab:results-polarimetric}
\begin{tabular}{lccc}
\toprule
\textbf{Model} & \scalebox{0.95}{\textbf{Acquire Rate}} & \textbf{EER} & \scalebox{0.95}{\textbf{TAR@FAR=1\%}} \\
\midrule
        \rowcolor[HTML]{eafaf1}
         \multicolumn{4}{c}{\textbf{Open source MLLMs}} 
        \\
Qwen3-VL-8B-\scalebox{0.95}{Instruct}            & 100\% & 12.90\% & 25.40\% \\
InternVL3-9B                    & 100\% & 27.56\% & 13.86\% \\
Gemma-3-4B-it                   & 100\% & 35.70\% & 0.00\%  \\
\midrule
        \rowcolor[HTML]{FFF7ED}
         \multicolumn{4}{c}{\textbf{Face Recognition Models}} 
        \\
xEdgeFace                       & 100\% & 1.60\%  & 97.70\% \\
EdgeFace                        & 100\% & 22.80\% & 33.60\% \\
\bottomrule
\end{tabular}
\end{table}

\begin{table*}[t]
\centering
\setlength{\tabcolsep}{6.5pt}
\caption{In-context learning using Qwen3-VL-8B-Instruct across modalities in the MCXFace dataset. AR denotes Acquire Rate. TAR is reported at FAR=1\%. }
\label{tab:results-mcxface-icl}
\begin{tabular}{l|ccc|ccc|ccc}
\toprule
 & \multicolumn{9}{c}{\textbf{Modality}} \\
\cmidrule(lr){2-10}
\textbf{No. Examples}
& \multicolumn{3}{c}{VIS--NIR}
& \multicolumn{3}{c}{VIS--SWIR}
& \multicolumn{3}{c}{VIS--THERMAL} \\
\cmidrule(lr){2-4}\cmidrule(lr){5-7}\cmidrule(lr){8-10}
 & AR$\uparrow$ & EER$\downarrow$  & TAR$\uparrow$ 
 & AR$\uparrow$ & EER$\downarrow$  & TAR$\uparrow$ 
 & AR$\uparrow$ & EER$\downarrow$  & TAR$\uparrow$  \\
\midrule
0 (Zero-shot)  & 100\% & \textbf{2.90}\% & \textbf{85.70}\% & 100\% & \textbf{12.10}\% & \textbf{13.00\%} & 100\% & \textbf{28.80\%} & 4.00\% \\ 
\midrule
1 positive + 1 negative & 100\% & 8.80\% & 72.00\% & 100\% & 28.20\% & 4.50\% & 100\% & \textbf{28.80\%} & \textbf{13.20\%} \\

\bottomrule
\end{tabular}
\end{table*}

\subsubsection{Baselines}
As baselines in our experiments, we compare the performance of MLLMs with a heterogeneous face recognition model (xEdgeFace) for different modalities and also a classical face recognition model (EdgeFace). These two baseline models have similar network structure and are different in training.

\paragraph{EdgeFace:} EdgeFace \cite{george2023edgeface} is a state-of-the-art light-weight face recognition model based on 
CNN-Transformer  backbone. This model is trained on RGB images from the WebFace dataset \cite{zhu2021webface260m}, and we do not further finetune it for other modalities in our HFR experiments. Our goal is to use this model as a baseline to see how a model trained for visual (RGB) images perform for HFR (for comparing with MLLMs in zero-shot setup). While EdgeFace is a lightweight face recognition model, it is amongst top-performing models on different face recognition benchmarks.

\paragraph{xEdgeFace:} xEdgeFace \cite{xedgeface} is a contrastive self-distillation framework for adapting a pretrained face recognition backbone to heterogeneous face recognition. The method retains the original EdgeFace \cite{george2023edgeface} architecture and introduces cross-modal alignment together with teacher–student supervision to enable effective adaptation while preserving performance on the source (RGB) face recognition task. As a result, the model acquires robust heterogeneous recognition capability without catastrophic forgetting. Despite its lightweight design, xEdgeFace achieves performance comparable to or surpassing state-of-the-art methods on multiple HFR benchmarks.

\subsubsection{Implementation Details}
We ran our experiments on a system equipped with NVIDIA H100.  We used the vLLM repository\footnote{https://github.com/vllm-project/vllm} to implement our evaluation of MLLM for HFR.
The source code of our experiments is publicly available\footnote{\href{https://gitlab.idiap.ch/biometric/code.hfr_mllm}{https://gitlab.idiap.ch/biometric/code.hfr\_mllm}.}.

\subsection{Experimental Results}
\paragraph{Performance on MCXFace Dataset:}
Table~\ref{tab:results-mcxface} reports the performance of multiple open-source MLLMs for HFR under VIS-NIR, VIS-SWIR, and VIS-THERMAL modalities from the MCXFace dataset. 
The table also compares the performance of MLLMs with pretrained face recognition (EdgeFace) and HFR (xEdgeFace) models. 
For the performance of each model, we report Acquire Rate (AR), Equal Error Rate (EER), and True Accept Rate (TAR) at FAR of 1\%. 
While the majority of MLLMs have AR of 100\%, in some cases (for Mistral-3.1 and Aya-Vision), the model was not able to provide similarity score for some samples. 

For all modalities in Table~\ref{tab:results-mcxface}, we observe that the pretrained HFR model (xEdgeFace) achieves better results than MLLMs. In most tasks, the pretrained face recognition (EdgeFace) also has a better performance than MLLMs, except for VIS-THERMAL, where Qwen3-VL-4B-Instruct gets a better EER value than EdgeFace. However,  for VIS-THERMAL, EdgeFace has a better TAR value.
Among different modalities, VIS-NIR appears to be the easiest for MLLMs and VIS-THERMAL is the hardest task. This can be due to the fact that MLLMs have been mainly trained on visual (RGB)  images. While NIR images are similar to gray images, thermal images look substantially different. Meanwhile, we should note that VIS-THERMAL is also the most difficult modality for pretrained face recognition (EdgeFace) and HFR (xEdgeFace) models.

In the remaining experiments, we focus on top-performing MLLMs in Table~\ref{tab:results-mcxface} and further evaluate their performance for HFR. 

\paragraph{Performance on CASIA Dataset:}
Table~\ref{tab:results-casia} compares the performance of MLLMs for VIS-NIR modality with baselines (xEdgeFace and EdgeFace) on the CASIA dataset. Similar to VIS-NIR evaluation on the MCXFace dataset (Table~\ref{tab:results-mcxface}), we can see that MLLMs have inferior performance in terms of EER and TAR compared to baselines (xEdgeFace and EdgeFace) on the CASIA dataset. Comparing the values with Table~\ref{tab:results-mcxface}, Qwen3-VL has same EER but much lower TAR value. Gemma-3 also has inferior performance in terms of EER and TAR. In contrast, InternVL3 has worse EER but slightly better TAR value that was achieved on the MCXFace dataset.

\paragraph{Performance on Polarimetric Dataset:}
Table~\ref{tab:results-polarimetric} also reports the performance of MLLMs for VIS-THERMAL heterogeneous face recognition on the Polarimetric dataset and compares with baselines (xEdgeFace and EdgeFace). As can be seen, thermal images are difficult for MLLMs to match, which makes VIS-THERMAL heterogeneous face recognition a challenging task. 
Compared to Table~\ref{tab:results-mcxface}, all MLLMs achieved better EER values and  better TAR (except for Gemma-3 which has slightly lower TAR). 
Similar to Table~\ref{tab:results-mcxface}, the pretrained HFR model (xEdgeFace) has the best performance. While Qwen3-VL achieves  better EER than EdgeFace, still EdgeFace performs better in terms of TAR.

\paragraph{In-Context Learning:}
As another experiment, we evaluate MLLMs under in-context learning setup, in which we provide example of images from both modalities before asking the MLLM to provide similarity score for test images.
The new system prompt is as follows: 
\textit{``You are an AI assistant specialised in heterogeneous face recognition. I give you two face images taken under \texttt{modality\_1} and \texttt{modality\_2}. On a scale from 0 to 100, how likely (as a single number) are these two faces of the same person? Only output a single number (no other text)."}
Then we provide examples of images from different modalities but from the same subject with the following prompt: 
\textit{``Here is an example of two images of the **same** person taken under \texttt{modality\_1} and \texttt{modality\_2}. You need to output a high similarity score for such pairs."}
Similarly, we provide examples of images from different modalities and from different subjects with the following prompt; 
\textit{``Here is an example of two images of **different** people taken under \texttt{modality\_1} and \texttt{modality\_2}. You need to output a low similarity score for such pairs."} Table~\ref{tab:results-mcxface-icl} compares the performance of Qwen3-VL-8B-Instruct in zero-shot setting versus in-context learning with one positive pair and one negative pair examples. While in-context learning provides more information through example, it has not improved the performance of the model. This can indicate the difficulty of HFR which requires matching faces based on fine-grained details in two images from different modalities to distinguish the identity in the open-set evaluation.

\section{Conclusion}\label{sec:conclusion}
In this paper, we explored the application of  MLLMs for heterogeneous face recognition. Heterogeneous face recognition requires comparison of face images from two different modalities. 
Multimodal large language models are often trained on general datasets, including RGB images. Although MLLMs have achieved remarkable performance on different downstream computer vision tasks, they are primarily trained on general-purpose data dominated by RGB imagery. Our experimental results show that,  despite their recent advances, current MLLMs struggle to match information from different image modalities for heterogeneous face recognition, including VIS–NIR, VIS–SWIR, and VIS–THERMAL scenarios. In comparison to classical face recognition systems specifically developed for face biometric, MLLMs have substantial performance gaps, particularly under challenging cross-spectral conditions. 
We also investigated in-context learning by showing MLLM example of positive and negative pairs of images from different modalities. However, our findings indicate such examples could not be sufficient for MLLMs to enhance their understanding of given images and distinguish two face images from different modalities.  
Overall, our study highlights the current limitations of MLLMs for heterogeneous face recognition and shows the importance of modality-aware training and also rigorous evaluation of the models for biometrics. While MLLMs hold promise as general-purpose multimodal models, significant advances are required before they can be reliably applied to sensitive biometric recognition tasks.

\section*{Ethics Statement}
This work investigates the application of MLLMs for heterogeneous face recognition, which is a biometric task with clear ethical, legal, and societal implications. Generally, face recognition models can pose risks related to privacy, surveillance, and misuse if deployed without sufficient safeguards. Our study is conducted with the goal of understanding potentials and limitations of MLLMs when used for HFR.
Importantly, this paper does not propose a deployable face recognition system, nor does it advocate the use of MLLMs for real-world biometric authentication or surveillance. On the contrary, our results demonstrate that current MLLMs exhibit significant performance gaps and unreliability in cross-spectral face recognition scenarios. 
All our experiments are performed on existing publicly available heterogeneous face datasets collected under established research protocols. We also used open-source MLLMs in our study and excluded commercial models, ensuring that biometric data in our evaluation is not transmitted to a third party. 
Overall, this work contributes to the responsible development of biometric systems by providing an  assessment of MLLMs for HFR and by highlighting the necessity of rigorous evaluation before considering their use in sensitive identity-related applications.

\section*{Acknowledgment}
This work was funded by the European Union project CarMen (Grant Agreement No. 101168325).
	
{
    \small
    \bibliographystyle{ieeenat_fullname}
    \bibliography{main}
}


\end{document}